\documentclass{esub2acm}
\input epsf.tex

\newtheorem{theorem}{Theorem}[section]
\newtheorem{lemma}{Lemma}[section]
\newtheorem{assumption}{Assumption}[section]
\newtheorem{definition}{Definition}[section]
\newtheorem{observation}{Observation}[section]
\newtheorem{corollary}{Corollary}[section]
\begin{document}
\title{Robustness of Regional Matching Scheme over Global Matching Scheme} 
\author{Liang Chen and Naoyuki Tokuda\\
         Utsunomiya University, Utsunomiya, Japan}
\begin{abstract}
The paper has established and verified the theory prevailing widely among image and pattern recognition specialists that the bottom-up indirect regional matching process is the more stable and the more robust than the global matching process against concentrated types of noise represented by clutter, outlier or occlusion in the imagery. We have demonstrated this by analyzing the effect of concentrated noise on a typical decision making process of a simplified two candidate voting model where our theorem establishes the lower bounds to a critical breakdown point of election (or decision) result by the bottom-up matching process are greater than the exact bound of the global matching process implying that the former regional process is capable of  accommodating a higher level of noise than the latter global process before the result of decision overturns.

 We present a convincing experimental verification supporting not only the theory by a white-black flag recognition problem in the presence of localized noise but also the validity of the conjecture by a facial recognition problem that the theorem remains valid for other decision making processes involving an important dimension-reducing transform such as principal component analysis or a Gabor transform.

\end{abstract}


\begin{bottomstuff}
An extended abstract of the preliminary version of the paper has been presented at AIDA'99 (International ICSC Symposium on Advances in Intelligent Data Analysis),  Rochester, New York in June of 1999.

\begin{authinfo}
\address{Computer Science Department, Utsunomiya University, Utsunomiya, Japan 321-8505; e-mail: lchen@alfin.mine.utsunomiya-u.ac.jp, tokuda@cc.utsunomiya-u.ac.jp.}
\end{authinfo}


\end{bottomstuff}
\markboth{L. Chen and N. Tokuda}
     {Robustness of Regional Matching Scheme over Global Matching Scheme}
\maketitle

\section{Introduction}
Consider some decision making process $G$ which arises quite frequently in many scientific researches as well as in daily life events. Given several candidates (or selections) to choose, we must choose one candidate (selection) by using the voting scheme or by matching the features extracted from the entire area (or the nation). We call this $G$ as a global voting ( or matching) method. Many countries adopt the system in choosing a President of the country as a Peruvian president. Now let us convert $G$ to a new version $r(G)$, where we decide the winner of each pre-divided regions by $G$ but we make the final decision by a simple majority of the number of the winning regions by adopting the ``winners-take-all" principle within the pre-divided regions. The latter converted version $r(G)$ is called a regional voting ( or matching) method. A most typical regional voting method is the US presidential election system. It is the robustness of the decision making by $G$ or $r(G)$ processes that we want to clarify in the present analysis against a concentrated type of noise.

Here, $G$ could be any decision making procedure. In the simple voting system, $G$ makes use of voting and makes a decision by the majority of votes counted. In the facial recognition problem, for example, $G$ based on this voting system may involve a pixel-by-pixel comparison between the two facial images to be compared or we could use any of well-known dimension-reducing schemes such as the principal component analysis (PCA) \cite{jolliffe} or a Gabor transform \cite{TSLee} as $G$, the former PCA scheme leading to the famous eigenface method \cite{turkpentland}. The resulting reduced facial space of PCA is instrumental in obtaining the major eigenfaces of the training, for example and it is easy to find the co-ordinate of a new face projected into the facial space, and then make a recognition decision by matching the projection with stored images of models. Then the corresponding $r(G)$ could be implemented first by dividing the whole two dimensional picture of rectangle frame into smaller regions (of equal size in our analysis). Within each region we make use of the PCA method in decision making as $G$ and we make the final decision of the whole picture in accordance with the simple majority principle in the number of the winning regions where the winner gets all the votes of the region. It is the improved stability of $r(G)$ over that of $G$ that has motivated our current research as verified by several convincing numerical examples given in this paper. 

The purpose of the present paper is to elucidate and clarify a basic mechanism why the regional matching method has an advantage in stability over the global matching scheme against noise. A simplest voting model is selected for analysis where each cell in the nation consists of one vote, thus $G$ itself can be regarded as a simple national voting scheme where the winner is decided by a simple majority principle. To simplify the analysis on the regional matching, we divide the nation into smaller regions of equal size where the winner is decided by the number of winning regions, the winner of a region being decided by $G$. We set up a noise-and-voting model for this simple situation and show that when the size of the regions is reasonably small, the regional voting scheme is more stable than the national voting scheme. A conjecture is made that this model is valid in a more general decision making process where $G$ involves a decision making process by PCA matching or Gabor matching. We present a convincing experimental verification to support the conjecture in appendix.

The present paper is constructed as follows. In Section \ref{sec:theorems}, we first give precise definitions on noise, noise-concentrated area and the number of noise-contaminated regions including basic assumptions used in the analysis.  We prove Theorem \ref{th:1} which relates the noise-concentrated area and the number of potential noise-contaminated regions. Theorem \ref{th:2} shows how we can improve the relation. The resulting  Corollaries \ref{cor:1} and \ref{cor:2} corresponding to Theorems \ref{th:1} and \ref{th:2} respectively give the lower bounds of the  noise level to a breakdown point of decision beyond which the decision of the voting may overturn. In Section \ref{sec:conclusion-discussion}, we examine the results of Section \ref{sec:theorems} from various angles. A very convincing experimental verification of the theory is presented in Section \ref{sec:black-white} using a black-and-white flag recognition problem confirming the validity of the theory on a pixel-by-pixel basis. An experimental verification given in appendix supports the conjecture that the theory developed for the pixel-by-pixel voting process remains valid for more general decision making processes involving dimension reducing schemes such as PCA or Gabor transform.

\section{Theorems} \label{sec:theorems}

\subsection{Notations \label{subsec:theorems-notations}}
Important notations and basic assumptions used in the paper will be summarized here.

We suppose that the nation (or the entire image for an image application) consists of $N$ unit cells (or pixels) each having one vote to exercise; for simplicity the nation is always represented by a rectangle of size $l \times m$, so that $N = l \times m$. The nation on the other hand can be partitioned into $K$ square, equal sized regions of $m_{r}\times m_{r}$ each. We also assume that both $l$ and $m$ are divisible by $m_{r}$ and that the pair of the opposing edges along the outer boundary of the rectangular nation are to glide onto the other end as glued together so that the nation can be partitioned into a total of $m_{r}^{2}=N/K$  different partitions.
Like $m_{r}$  which is the length scale of a square region,  $m_{n}$ denotes that of a noise-concentrated block, with the terms ``noise" and ``noise-concentrated block" being defined in Definition 2 of next subsection.

\subsection{Main Theorems \label{subsec:theorems-maintheorems}}
We analyze in this paper a very simplified model allowing only two candidates in the election, say, two candidates {\it A} and {\it B}. Without losing generality, we assume, in the absence of external sources of noise, $A\%$ of total cells vote for {\it A} and $B\%$ cells vote for {\it B} so that 
\[A\%+B\%=1  \hspace{5mm} \mbox{and} \hspace{5mm} A\% > B\%. \]

We discuss our possible extension to an $n$ candidates system in section \ref{subsubsec:3candidates} We examine the effect of ``concentrated" noise on the decision making process of election results by the global voting and the regional voting. In image application, such noise is often observed in case the imagery contains transparency, specular reflections, shadows, fragmented occlusion as seen through branches of a tree or a sun shade and occlusion \cite{blackanandan}. 

The formal definition of concentrated noise as well as the formal definition of the global voting and regional voting is given below.

\begin{definition}[VOTING] 

\hspace{1cm}

$\bullet$ {\it National Voting--} The entire population $N$ of the nation vote either for Candidate {\em A} or {\em B} and Candidate {\em A} wins {\em if and only if} he gets a majority of the $N$ votes. 

$\bullet$ {\it Regional Voting--} The population $m_{r}^{2}$ ($=N/K$ for $K$ regions) of a region vote for Candidate {\em A} or {\em B} and a majority of votes determine the candidate of the region and a majority of the $K$ winning regions, not the majority of the entire population $N$ of the nation,  determines the winner for  the nation. 
\end{definition}

\begin{definition}[NOISE]

\hspace{1cm}

$\bullet$ We call a set of noise {\em anti-A-noise} (or {\em anti-B-noise}) if all the cells under influence will vote for {\em B} (or {\em A}) regardless of whether it originally votes for {\em A} or {\em B}. The number of the cells under influence is called the number of noise units.

$\bullet$ We call a vote noise-contaminated if the vote of a cell happens to undergo a change either from candidate {\em A} to {\em  B} or from candidate {\em B} to {\em  A}  under some changes of environmental conditions. The noise-contaminated vote undergoing a change from candidate {\em A} to {\em B} (or {\em B} to {\em A}) is especially called {\em anti-A-noise-contaminated vote} (or {\em anti-B-noise-contaminated vote}) respectively. 

$\bullet$ Anti-{\em A}-noise-concentrated blocks (Anti-{\em B}-noise-concentrated blocks) are defined as non-overlapped $m_{n} \times m_{n}$ sized areas among which all the cells are under influence of anti-{\em A}-noise (anti-{\em B}-noise). 

$\bullet$ The anti-{\em A}-noise-concentrated (anti-{\em B}-noise-concentrated) area is defined as the union of all anti-{\em A}-noise-concentrated (anti-{\em B}-noise-concentrated) blocks. 

$\bullet$ The region is defined to be anti-{\em A}-noise-contaminated (anti-{\em B}-noise-contaminated) {\em if and only if}  the conjunction set of the region and the anti-{\em A}-noise-concentrated (anti-{\em B}-noise-concentrated) area is not empty.

\end{definition}

In the analysis, we assume that there is only anti-{\em A}-noise. 

\begin{assumption} \label{asp:onlyA}
The effects of anti-{\em B}-noise on election results will be ignored in the analysis .
\end{assumption}

This assumption will be justified for the following two reasons. Firstly the anti-{\em B}-noise  and the anti-{\em A}-noise are independent so that we may consider the effect of the anti-{\em A}-noise entirely independent of the anti-{\em B}-noise. Secondly we want to establish a lower bound to a breakdown point in the prevailing situation of $A\% > B\%$. We see that the anti-{\em A}-noise gives a lower bound in terms of a noise level up to which we can accommodate before the results of the regional as well as the global voting reverse or overturn. The result for the regional voting will be established in Theorems \ref{th:1} and \ref{th:2} and Corollaries \ref{cor:1} and \ref{cor:2} while the exact bound for the national voting is given in Observation \ref{obs:1}. 

As we have emphasized, we only consider locally ``concentrated" noise. Thus we have:

\begin{assumption} \label{asp:concentrated-only}
All the anti-{\em A}-noise is within anti-{\em A}-noise-concentrated area.
\end{assumption}

This assumption will always hold, because we can regard a smallest block size of a single cell as the size of a noise-concentrated block at a worst case.  For general cases of  $m_{n} \times m_{n}$ cells excluding blocks consisting of single cell size, it is not difficult to conclude that a possible error between the anti-{\em A}-noise of the nation and the anti-{\em A}-noise-concentrated area becomes negligibly small as the size of each of the noise-influenced areas increases sufficiently. Thus  this assumption will not affect the validity of all the following theorems. We will clarify the situation in Observation \ref{obs:3} of section \ref{subsubsec:sizeofblock}.

The following assumption is made in terms of the definitions we introduce.

\begin{assumption} \label{asp:averagedistribution}

\hspace{1cm}

$\bullet$ {\it Average Distribution Assumption--} We assume that in the absence of noise, the voting distribution of the undisturbed national voting prevails in {\em any} sufficiently large size areas whether  consisting of a continuous part of the nation or of randomly chosen blocks of cells. 

$\bullet$ {\it Region Size--} We assume that the size of equally partitioned regions is sufficiently large so that in the absence of noise, the average distribution assumption above holds.
\end{assumption} 

The assumption implies that, in the absence of noise, the global voting behavior of {\em A\%} and {\em B\%} prevails in each of the regions such that there are almost {\em A\%(N/K)} cells voting for {\em A} and {\em B\%(N/K)} cells voting for {\em B}. This assumption can be relaxed (see section \ref{subsubsec:relaxing}). 

We conclude that, if Candidate {\em A} (or {\em B}) wins in the nation, so does Candidate {\em A} (or {\em B}) in each of the regions.

\begin{observation}  \label{obs:1}
If there exists more than $\frac {A\%-B\%}{2}\times N$ of anti-{\em A}-noise-contaminated votes, that is, if $\frac {A\%-50\%}{A\%}$ of the original votes cast for {\em A} should change to {\em B}, then the noise is effective in reversing the candidate selection from {\em A} to {\em B} in the national voting. We say the national voting can accommodate $\frac {A\%-B\%}{2}\times N$ noise before a reversal of the original voting result takes place.
\end{observation}

\begin{definition}
We call a region anti-{\em A}-noise-contaminated {\em if and only if}  the conjunction set of the region and the anti-{\em A}-noise-concentrated area is not empty.
\end{definition}

The following lemma shows that we can construct a partitioning of the nation such that the noise-concentrated blocks are concentrated into some fractions of all the regions. It gives a clue why the regional voting is capable of accommodating  a higher noise level than the averaged national voting because only a fraction of the entire regions absorb the dominant effects of the noise superimposed. 

\begin{lemma} \label{lem} 
For any given small positive integer $s$, we can always choose a partition of the rectangle nation into $K$ regions such that anti-{\em A}-noise is concentrated among a fractional $K'$ regions of the $K$ regions ($K'\leq K$) so that the total size of these $K'$ regions is less than that of the anti-{\em A}-concentrated area plus $s$ units.
\end{lemma}

\begin{proof} 
The lemma can be proved directly by considering the worst case: namely we can always divide the rectangle at worst case into  $K = l \times m$ regions of unit size, which means the above difference always vanishes. 
\end{proof}

Note that depending on the number of $s$ and the distribution of anti-{\em A}-contaminated noise, we do not always have to divide the nation into regions of unit size and a noise-concentrated block can be found, fulfilling the conclusion of the lemma.

The following theorems show the relations of the size of anti-{\em A}-noise-concentrated area and the total size of anti-{\em A}-noise-contaminated regions in the worst case which lead to the lower bounds to a breakdown point of decision.

\begin{theorem}  \label{th:1}
Let $S_{c}$ be the size of anti-{\em A}-noise-concentrated area and $S_{r}$ be the total size of anti-{\em A}-noise-contaminated regions of K-partitioned regional voting. We then have:
\begin{enumerate}
\item  $ \frac {S_{r}} {S_{c}} \leq  (\lceil \frac {m_{n}} {m_{r}} \rceil +1)^{2}\times { \frac {m_{r}^{2}} {m_{n}^{2}}}$.
\item  $S_{c} < \frac {m_{n}^{2}} {m_{r}^{2}} \times 1/(\lceil \frac {m_{n}} {m_{r}} \rceil +1)^{2} \times 50\% N$ is a  sufficient condition for the regional voting to retain the original candidate selection of  {\em A}. 
\end{enumerate}
\end{theorem}

\begin{proof}
Item 1 of the theorem follows immediately for ${m_{r}}>{m_{n}}$ because for any combination of partitioning, each of the anti-{\em A}-noise-concentrated block of size  $m_{n}\times m_{n}$ can at best  ``contaminate" $(\lceil \frac {m_{n}} {m_{r}} \rceil +1)^{2}$ equal size regions of $m_{r} \times m_{r}$. It is a simple matter to confirm that the conclusion is valid for ${m_{ n }} \geq {m_{ r }}$ as well. Item 2 of the theorem comes from the fact that at most $50\%$ of the $K$ regions can  be contaminated when $S_{c} < \frac {m_{n}^{2}} {m_{r}^{2}} \times 1/(\lceil \frac {m_{n}} {m_{r}} \rceil +1)^{2} \times 50\% N$.
\end{proof}

We immediately have the following Corollary.

\begin{corollary}  \label{cor:1}
The original candidate selection of {\it A} can accommodate at least $\frac {m_{n}^{2}} {m_{r}^{2}} \times 1/(\lceil \frac {m_{n}} {m_{r}} \rceil +1)^{2} \times (A\%/2)  N$ anti-{\it A}-noise  (i.e. $\frac {m_{n}^{2}} {m_{r}^{2}} \times 1/(\lceil \frac {m_{n}} {m_{r}} \rceil +1)^{2} \times 50\% $ of the cells voting for {\em A}), before the candidate selection is reversed.
\end{corollary}

Theorem \ref{th:1} and Corollary \ref{cor:1} show clearly that to retain the original candidate selection of {\em A} in the  regional voting, a larger subdivision of the nation namely into a smaller size region leads to a higher stability, provided that Assumption \ref{asp:averagedistribution} on region size remains valid.

\subsection{Further Improvement by Shifting Strategy}
The bounds of Theorem \ref{th:1} and Corollary \ref{cor:1} can be further improved by exploiting the {\it Shifting Strategy} of \cite{hochbaummaass}. We first define a {\it Shifting Strategy} for some operation $\Lambda$ with respect to a square region embedded within a rectangular nation.

{\em Shifting Strategy for Certain Action} $\Lambda$ \\
Consider the partitioning of a rectangular nation into $m_{r} \times m_{r}$
square regions where $m_{r}$ is some arbitrary integer. \\
Repeat step 1 to step 2 $m_{r}$ times:
\begin{enumerate}
\item Move all the vertical partition lines to right by one cell, repeat
step 2 for $m_{r}$  times;
 \item Move all the horizontal lines up by one cell, execute Action
$\Lambda$.
\end{enumerate}

The shifting strategy enumerates  all the possible different partitioning of the nation. Now by replacing  Action $\Lambda$ of the strategy with the regional voting subject to the same noise environment, we show by Theorem \ref{th:2} below how we can improve Theorem \ref{th:1}.

\begin{theorem} \label{th:2}
Under the assumptions of Theorem \ref{th:1}, the shifting strategy ensures that there exists at least one partition satisfying the following properties:
\begin{enumerate}
\item  $\frac{S_{r}}{S_{c}} \leq  (\frac {m_{r}+m_{n}-1}{m_{n}})^{2}$. 
\item The sufficient condition for the regional voting to retain the original candidate selection of  {\em A} can be improved to: $S_{c} < (\frac {m_{n}}{m_{r} + m_{n}-1})^{2} \times 50\% N$.
\end{enumerate}
\end{theorem}

\begin{proof}
To prove item 1, we must show that among all the possible $m_{r}^{2}$ different partitions that the Shifting Strategy can possibly generate, each of $m_{n} \times m_{n}$ size anti-{\em A}-noise-concentrated block is capable of contaminating the total  of $(m_{n}+m_{r}-1)^{2}$ different regions. Once this is done, the Pigeon Hole Principle \cite{ahoullman} ensures that there is at least one partition in which the existing $S_{c}/m_{n}^{2}$ anti-{\em A}-noise-concentrated blocks contaminate at most $(S_{c}/m_{n}^{2})(m_{n}-m_{r}+1)^{2} / m_{r}^{2}$ regions. 

We prove this for $m_{r} \geq m_{n}$ first. Among all possible  $m_{r}^{2}$ partitions, $(m_{n}-1)^{2}$ partitions divide the block into 4 regions, $2 \times (m_{n}-1)+2 \times (m_{n}-1)(m_{r}-m_{n})$ of them divide the block into 2 regions, while $(m_{r}-m_{n}+1)^{2}$ of the partitions can not divide the block into more than one region. Summing them up, the noise-concentrated block is divided into $(m_{n}+m_{r}-1)^{2}$ different regions. Figure 1 illustrates the three cases above for $m_{n}=5$ and $m_{r}=8$, for example. We see that the partitioning through the solid dots of the figure divide the noise block into 4 regions, those through the crossed points into 2 regions, while those through the hollow dots may not be able to divide the block. 

For $m_{r} < m_{n}$, we enumerate each of all the $m_{r} \times m_{r}$ possible partitions. We know that the block would be divided into $\lceil \frac{m_{n}}{m_{r}} \rceil \cdot \lceil \frac{m_{n}}{m_{r}} \rceil$, $\lceil \frac{m_{n}}{m_{r}} \rceil \cdot (\lceil \frac{m_{n}-1}{m_{r}} \rceil+1)$, $\lceil \frac{m_{n}}{m_{r}} \rceil \cdot (\lceil \frac{m_{n}-2}{m_{r}} \rceil+1)$, $\lceil \frac{m_{n}}{m_{r}} \rceil \cdot (\lceil \frac{m_{n}-3}{m_{r}} \rceil+1)$,  $\cdots$, $\lceil \frac{m_{n}}{m_{r}} \rceil \cdot (\lceil \frac{m_{n}-m_{r}+1}{m_{r}} \rceil+1)$;
$(\lceil \frac{m_{n}-1}{m_{r}} \rceil+1) \cdot \lceil \frac{m_{n}}{m_{r}} \rceil$, $(\lceil \frac{m_{n}-1}{m_{r}} \rceil+1) \cdot (\lceil \frac{m_{n}-1}{m_{r}} \rceil+1)$, $(\lceil \frac{m_{n}-1}{m_{r}} \rceil+1) \cdot (\lceil \frac{m_{n}-2}{m_{r}} \rceil+1)$, $(\lceil \frac{m_{n}-1}{m_{r}} \rceil+1) \cdot (\lceil \frac{m_{n}-3}{m_{r}} \rceil+1)$,  $\cdots$, $(\lceil \frac{m_{n}-1}{m_{r}} \rceil+1) \cdot (\lceil \frac{m_{n}-m_{r}+1}{m_{r}} \rceil+1)$;
$(\lceil \frac{m_{n}-2}{m_{r}} \rceil+1) \cdot \lceil \frac{m_{n}}{m_{r}} \rceil$, $(\lceil \frac{m_{n}-2}{m_{r}} \rceil+1) \cdot (\lceil \frac{m_{n}-1}{m_{r}} \rceil+1)$, $(\lceil \frac{m_{n}-2}{m_{r}} \rceil+1) \cdot (\lceil \frac{m_{n}-2}{m_{r}} \rceil+1)$, $(\lceil \frac{m_{n}-2}{m_{r}} \rceil+1) \cdot (\lceil \frac{m_{n}-3}{m_{r}} \rceil+1)$,  $\cdots$, $(\lceil \frac{m_{n}-2}{m_{r}} \rceil+1) \cdot (\lceil \frac{m_{n}-m_{r}+1}{m_{r}} \rceil+1)$;
$\cdots$;
$(\lceil \frac{m_{n}-m_{r}+1}{m_{r}} \rceil+1) \cdot \lceil \frac{m_{n}}{m_{r}} \rceil$, $(\lceil \frac{m_{n}-m_{r}+1}{m_{r}} \rceil+1) \cdot (\lceil \frac{m_{n}-1}{m_{r}} \rceil+1)$, $(\lceil \frac{m_{n}-m_{r}+1}{m_{r}} \rceil+1) \cdot (\lceil \frac{m_{n}-2}{m_{r}} \rceil+1)$, $(\lceil \frac{m_{n}-m_{r}+1}{m_{r}} \rceil+1) \cdot (\lceil \frac{m_{n}-3}{m_{r}} \rceil+1)$,  $\cdots$, $(\lceil \frac{m_{n}-m_{r}+1}{m_{r}} \rceil+1) \cdot (\lceil \frac{m_{n}-m_{r}+1}{m_{r}} \rceil+1)$ regions separately. Summing up all the possible terms above by means of  formula 
\[ \sum_{i=0}^{m_{r}-1} \lceil \frac{m_{n}-i}{m_{r}} \rceil = m_{n}, \]
we know that the block will be divided into $(m_{n}-m_{r}+1)^{2}$ different regions for all the $m_{r} \times m_{r}$ different partitions. 

\begin{figure}
\centerline {\epsfxsize=50 mm \epsfbox{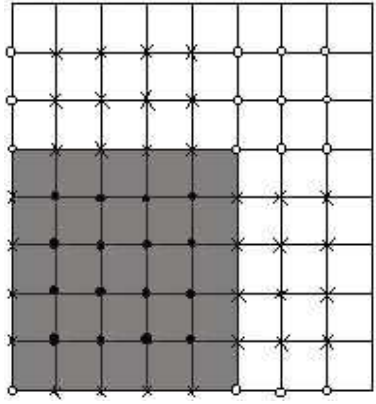}}
\caption{Different Partitioning in the Nation}
\end{figure}

Item 2 follows from item 1 since the condition that the number of noise-contaminated regions is to be less than 50\% of the total number of regions constitutes a sufficient condition for the original candidate selection of ${A}$.
\end{proof} 

\begin{corollary} \label{cor:2}
For a fixed sized noise-contaminated region and a fixed sized, equally partitioned  region, we can find at least one partition such that a specific partition can accommodate at least: $(\frac {m_{n}}{m_{r} + m_{n}-1})^{2} \times (A\%/2) N$ anti-{\em A}-noise-contaminated votes (that is $(\frac {m_{n}}{m_{r} + m_{n}-1})^{2} \times 50\%$ of all the votes originally given to {\em A}) before the result of candidate selection is reversed.
\end{corollary} 

{\it CONJECTURE:
Theorem \ref{th:1} and Theorem \ref{th:2} and related Corollaries  remain valid for a more general $G$ including the features matching by PCA analysis.}

We confirm this conjecture by means of experimental verifications in Appendix.

\section{Conclusion and Discussion} \label{sec:conclusion-discussion}

The detailed analysis on the regional and national voting convincingly shows that the regional voting is the more stable and robust of the two. This is in agreement with our physical intuition as supported by lemma \ref{lem}, that only a small fraction of the entire regions absorb the heavily concentrated effects of noise. We would like to give some concrete examples numerically below where possible.

\subsection{Conclusion \label{subsec:conclusion}}

\subsubsection{Robustness of Regional Voting}

Table \ref{tb:1} is computed from the formula of Corollary \ref{cor:1} and Observation \ref{obs:1} of Section \ref{subsec:theorems-maintheorems} for several values of $(A\%-B\%)$.

\begin{table*}[hpt]
\caption{Stability Margins of Regional Voting and National Voting for N=10000 \label{tb:1}}
\begin{quote}
\em Stability Margins of anti-A-noise-contaminated votes which Regional and National voting can accommodate before the decision reverses. 
\end{quote}
\begin{center}
{
\small
\begin{tabular}{{|c|c|c|c|}} \hline
&\multicolumn{2}{c|}{Regional Voting}& \\ \cline{2-3}
$A\%-B\%$&$m_{n}/m_{r}=1$&$m_{n}/m_{r}=2$&National Voting \\ \hline
5\%&656&1167&250\\ \hline
10\%&688&1222&500\\ \hline
20\%&750&1333&1000\\ \hline
\end{tabular}
}
\end{center}
\end{table*}

We see that the regional voting is always more stable and robust than the national voting when $A\%-B\%$ is not too large, say 5\% and 10\% and this robustness increases as $\frac {m_{n}} {m_{r}}$ increases or the partitioned region size becomes smaller.

For larger  values of $A\%-B\%$, say, 20\%, at $m_{n}/m_{r}=1$, the lower bound of regional voting given by Corollary \ref{cor:1} is smaller than the exact bound of national voting. We still believe that the regional voting can still be more stable than the national voting, because in counting the number of possible ``losing" regions in Theorem \ref{th:1}, we have counted the number of all the noise-contaminated regions including those which  still have a margin to the breakdown point retaining the pro-{\it A} region.
In fact, we have excluded only those regions entirely {\em clean or free of any noise}.  

\subsubsection{Improvement by Shifting Strategy}

Given some distributed noise-concentrated area as the union of all noise-concentrated blocks, Theorem \ref{th:2} and Corollary \ref{cor:2} provide a method of improving the stability of the regional matching as shown in Table \ref{tb:2}. A larger improvement is evident for smaller $A\%-B\%$. 

\begin{table*}[hp]
\caption{Improved Regional Voting by Shifting Strategy Calculated for $N=10000$ \label{tb:2}}
\begin{center}
{
\small
\begin{tabular}{{|c|c|c|c|c|}} \hline
&\multicolumn{2}{c|}{$m_{n}=3,m_{r}=3$}&\multicolumn{2}{c|}{$m_{n}=4,m_{r}=2$}\\ \cline{2-5}
$A\%-B\%$&Corollary \ref{cor:1}& Corollary \ref{cor:2}& Corollary \ref{cor:1}& Corollary \ref{cor:2} \\ \hline
5\%&656&945&1167&1680\\ \hline
10\%&688&990&1222&1760\\ \hline
15\%&719&1035&1278&1840\\ \hline
20\%&750&1080&1333&1920\\ \hline
\end{tabular}
}
\end{center}
\end{table*}

\subsubsection{A Tradeoff on Region Size \label{subsubsec:tradeoff}}
By examining the theorems, we see that there must be a tradeoff for the size $m_{r}$ of the partitioned regions; the smaller size of a partitioned region increases the robustness of the regional voting but, Assumption \ref{asp:averagedistribution} requires the size not too small. This is remarkably well born out experimentally in figure 6 where at low as well as at high noise level, the recognition rate falls off if the entire image of ${80 \times 120}$ pixels is divided into more than 384 regions(a region size corresponding to $5 \times 5$ pixels) to which we come back later on.

\subsection{Discussion}

\subsubsection{What if there are more than three candidates? \label{subsubsec:3candidates}}
If the number of candidates exceeds two, the number of decision making processes increases. For example, we may allow each region to select top two or more candidates at a time, then make the candidate selection based on the summed results of all the regions. 
Here, we set up one simple model where the basic decision making principle adopted in the two candidate system is retained. Each of the regions selects only one (1) candidate according to a simple majority principle, and then the regional voting selects one candidate who wins a majority of the winning regions. Suppose there are candidates {\em A}, {\em B}, {\em C}, $\cdots$, and $A\% > B\% >C\% > \cdots$. The anti-A-noise is defined to convert the votes originally for {\em A} to {\em B} and keep other votes unchanged. We have the following theorem by exactly same proof of Theorem \ref{th:1}.

\begin{theorem} \label{th:3}

\hspace{1cm}\\
{\it 1.} The national voting can only accommodate at most $(A\%-B\%)/2\cdot N$ anti-{\it A}-noise-contaminated-votes, i.e. $(A\%-B\%)/2A\%$ among all the votes originally for {\it A}.\\
{\it 2.} Regional voting can accommodate at least 
$\frac {m_{n}^{2}} {m_{r}^{2}} \times 1/(\lceil \frac {m_{n}} {m_{r}} \rceil +1)^{2} \times (A\%/2)  N$
 anti-{\it A}-noise-contaminated votes 
i.e. $\frac {m_{n}^{2}} {m_{r}^{2}} \times 1/(\lceil \frac {m_{n}} {m_{r}} \rceil +1)^{2} \times 50\% $ among all the votes originally for {\it A}.
\end{theorem} 

We have an entirely same conclusion as in the two candidate system, confirming that the regional voting still accommodates a higher level of noise when $A\%$ and $B\%$ are very close.

\subsubsection{Effect of salt-and-pepper noise}
In sharp contrast to localized and thus concentrated noises we have assumed
in the present paper, we examine the effects of  impulse-type salt-and-pepper noise  (\cite{pratt}).

\begin{definition}
The white noise is a set of noise ``uniformly" distributed over the nation in such a way that in each of reasonably large sized areas whether composed of a
continuous  part of the nation or of randomly chosen blocks of cells, we
have a same percentage of noise.
\end{definition}

Consider only the impulse-type well dispersed anti-A-noise. The similar conclusion about the noise bound as Observation \ref{obs:1} can be obtained:
\begin{observation} \label{obs:2}
The global voting and regional voting can accommodate the same percentage of
salt-and-pepper noise.
\end{observation}
The observation could easily be proved by noting that in each region or the
whole nation, the candidate selection will not reverse unless there is less
than $\frac{1}{2}(A-B)\%$ of salt-and-pepper noise.

The observation shows that as long as the partitioned region is large enough
to allow the original distribution of the entire nation to prevail, we
expect no difference between the two decision systems in the presence of
salt-and-pepper noise. It is when the uniform distribution assumption fails between
the national voting and the regional voting as we have assumed in the paper
that the difference matters.

\subsubsection{Size of noise-concentrated blocks \label{subsubsec:sizeofblock}}

The generality of our analysis will heavily depends on the fact that a possible error between the total size of all noise generated and the union of all
non-overlapped noise-concentrated blocks be kept negligibly small. Let us cut out some $m_{n} \times m_{n}$ blocks out of reasonably concentrated noise-contaminated area. It is reasonable to assume that $m_{n} \times m_{n}$  is small compared to the size of any continuous parts of noise-contaminated area. We expect that a measure of what is left after cutting out blocks is surely quite small compared with the total number of noise. We demonstrate this formally below by showing why our theorems and corollaries should remain valid.

\begin{definition}
A line segment is called an orthodiameter of a continuos area of the nation, {\em if and only if} 
\begin{enumerate}
\item all the points of the segment lie within the area;
\item only the two end points of the line segment lie on the boundary of the area;
\item the line segment is parallel to the horizontal lines or the vertical lines comprising the boundary of the nation.
\end{enumerate}
\end{definition}

\begin{definition}
The orthomeasure of an area (continuous or detached) of the nation is defined by the length of the shortest orthodiameter of any continuous part of the area.
\end{definition}

\begin{observation} \label{obs:3}
Let us cut out as many $m_{n} \times m_{n}$ blocks as possible and let $S_{c}$ be the total size of all these blocks (i.e. the size of noise-concentrated area discussed in definition 2). Suppose {\em OM} be the orthomeasure of the set of noise influenced votes of the nation, and $N_{n}$ be the number of noise (or, the number of noise-affected cells). We have:
\[ \lim_{\frac{OM}{m_{n}-1} \rightarrow \infty} \frac{N_{n}-S_{c}}{N_{n}}=\lim_{\frac{OM}{m_{n}-1} \rightarrow \infty} \frac{N_{n}-S_{c}}{S_{c}}=0. \]
\end{observation}

Note that the above observation includes the situation that $N_{n}-S_{c}=0$ when $m_{n}=1$.

\subsubsection{All of the anti-{\em A}-noise-concentrated blocks are effective in reversing the votes ?}

Suppose that in a noise-concentrated area, only $r \%$ of votes for {\em A}  undergoing changes to {\em B} where $r$ is some constant within $[0,1]$ and  closer to 1. All the results in  theorems \ref{th:1}-\ref{th:2} remain the same while those  of corollaries \ref{cor:1}-\ref{cor:2} may be divided by $r$. We believe that the conclusion of section \ref{subsec:conclusion} still holds because in all the corollaries we always use quite much larger values for  lower boundaries as we have thrown away all the regions contaminated with a single noise.

\subsubsection{Relaxing the Average Distribution Assumption \label{subsubsec:relaxing}}

In view of the proofs of Theorems \ref{th:1}-\ref{th:2} given in Section \ref{sec:theorems}, the restriction of regional size in Assumption \ref{asp:averagedistribution} in Section \ref{subsec:theorems-maintheorems} is too strict and can be relaxed as follows. 

\begin{observation} \label{obs:4}
All the conclusions of Theorems \ref{th:1} and \ref{th:2} and Corollaries \ref{cor:1} and \ref{cor:2} still hold, as long as we can choose the size of partitioned regions large enough such that the voting distributions in the absence of noise for Candidates {\em A} and {\em B} in each of the regions  satisfy \[ A\% > B \%.\] 
\end{observation}
 We emphasize that the voting distributions in the regions do not have to follow that of the nation in the absence of noise. 

\section{Experiment: White or Black dominated Flags} \label{sec:black-white}
The first  example relates to a white-black mixed flag which we want to recognize either as  a white or a black dominated flag (see figure 2 for illustration where the cells in the figure denote a smallest unit of a ``pixel"). Unlike the second example to follow, this example applies the present theory directly on a pixel by pixel basis without resorting to features extracting transformation such as Turk \& Pentland method. 
The size of the ``nation" is $15 \times 24$ cells and the partitioned region comprises a $3 \times 3$  or a $4 \times 5$  block  which is  not either too large nor too small relative to the size of the nation. Suppose that a white-dominated flag is given as in figure 2-(1). This is confirmed easily by both global and regional voting because in the global voting, ``White" gets  207   votes while  ``Black"  153 votes; by regional vote counting based on a $4 \times 5$ regional partitioning, ``White" wins in  12 regions while ``Black" does in 4 regions, and within another 2 regions ``White" and ``Black" get same votes. If we further divide the nation into $3 \times 3$ sized regions. ``White" wins in  28 regions while ``Black" does in 12 regions in regional voting. Now we choose arbitrary $7$ pixels of figure 2-(1) randomly and introduce anti-White-noise blocks of a $5 \times 5$ areas so that every ``white" pixel within the block are transformed to a Black pixel with the probability of  ``0.7". As a result, 35 ``white" pixels are changed to ``Black" (are anti-white-noise contaminated) transforming figure 2-(1) to figure 2-(2).
By counting, we see that after the noise is added, the global voting will reverse the results of the candidate selection from ``White" to ``Black" dominated because this time ``Black" gets 188 votes while ``White" gets only 172 votes in global voting. But, by regional voting having the size of $4 \times 5$ cells, the original selection of ``White" dominated still remains valid, because this time ``White" wins in  10 regions while ``Black" does so in 6, and within another 2 regions ``White" and ``Black" get same votes. If we further divide the nation into $3 \times 3$ sized regions, we will see that ``White" wins in  25 regions while ``Black" does in 15 regions in regional voting increasing the stability margin thus confirming our theory.

\begin{figure*}[ht]
\centerline{\epsfxsize=70 mm \epsfbox{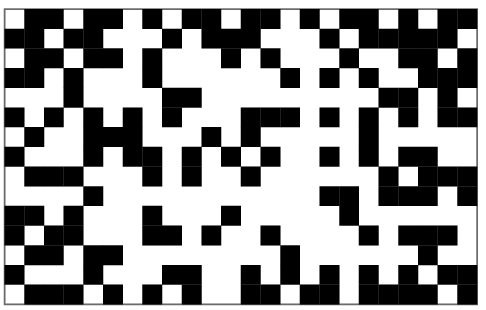} \epsfxsize=70 mm \epsfbox{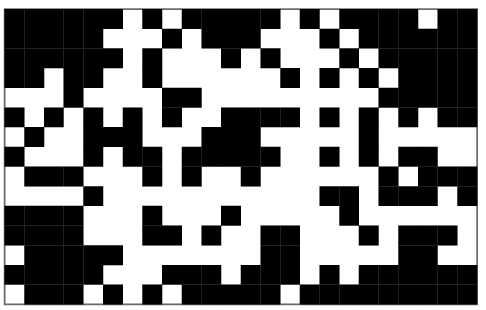}}
\centerline{(1) \hspace{68mm} (2)}
\caption{``White" or ``Black" Dominated?}
\end{figure*}

\section*{Appendix: Facial recognition}

We show a most convincing verification of the conjecture in practical image processing applications by carrying out a set of facial recognition experiments subject to localized concentrated noise. We have used the images of 16 people. As shown in figure 3-(2) and -(3), we introduce circular blocks of noise of Photo-shops version 4.0 randomly into the test images at a low and high noise levels of 25\% and 50\% levels(see figure 3). 
 \begin{figure*}[hp]
\centerline{\epsfxsize=112.9 mm \epsfbox{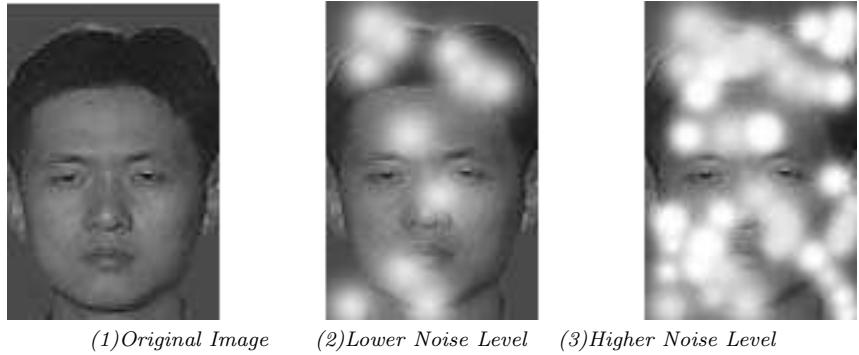}}
\centerline{\em (1)Original Image \hspace{0.4cm} (2)Lower Noise Level \hspace{0.2 cm} (3)Higher Noise Level}
\caption{ Typical Noise Contaminated Images At low and High Noise Level}
\end{figure*}
Circular blocks of noise  by  Photoshops have a density variation within the area ranging from 0 to 255 level. We have defined the area to be noise-affected if the density levels of the original image and those of the noise-affected image differ by 64 (i.e. $25\% \times 256$) in pixel density. The noise level in each of the images is shown in Table \ref{tb:3}. Each picture is of size $80 \times 120=9600$ pixels, the first row of each level indicates the numbers of noise, the second row the percentage of the noise.

\begin{table*}[hp]
\caption{The Noise of Each Images \label{tb:3}}
\begin{quote}
\em The first row of each level indicates the numbers of noise while the second gives he level of noise in percentage 
\end{quote}
\begin{center}
{
\small
\begin{tabular}{{|c|c|}*{16}{@{\hspace{0.02 cm}}c@{\hspace{0.02 cm}}|}} \hline
\multicolumn{2}{|@{\hspace{0.0 cm}}c@{\hspace{0.0 cm}}}{Noise suffered} & \multicolumn{16}{|c|}{Training face images of individuals (80 $\times$ 120 size)}\\ \cline{3-18}
\multicolumn{2}{|c|}{face images} & No.0 & No.1 & No.2 & No.3 & No.4 & No.5&No.6&No.7&No.8&No.9&No.10&No.11&No.12&No.13&No.14&No.15\\ \hline
Lower&No.&2757&2450&2429&1871&2004&1729&1959&2743&2502&2379&2043&2437&1566&2895&2517&1946 \\ \cline{2-18}
level &\%&28.7&25.5&25.3&19.5&20.9&18.0&20.4&28.6&26.1&24.8&21.3&25.4&16.3&30.2&26.2&20.3 \\ \hline
Higher&No.&5748&5474&5701&4372&4909&5051&5359&4464&4907&5073&4755&4900&4762&5377&5711&4811 \\ \cline{2-18}
lever&\%&59.9&57.0&59.4&45.5&51.1&52.6&55.8&46.5&51.1&52.8&49.5&51.0&49.6&56.0&59.5&50.1 \\ \hline
\end{tabular}
}
\end{center}
\end{table*}

Turk and Pentland's eigenvector algorithm \cite{turkpentland} is used for features extraction or data compression purpose where the eigenvector transformation operates uniformly not only over each of discrete pixels in the nation but also over  each of pixels of partitioned regions. Now that the effects of random noise remain random on the transformed planes without being magnified or filtered, we assume that the effects of noise blocks remain transparent to the transformation, implying that the conjecture remains valid. For convenience, we now resort to a new partition notation dividing the nation 1 (namely global), to 8, 16, 32, 64, $\cdots$, regions so that the division of 1 corresponds now to the original national matching. 

Recognition rates of the experiments are compared in figures 4, 5 and 6 illustrating the results of 1-region up to 16-, 32-, 64-, $\cdots$ regional matching at both low and high noise levels respectively. Figures 4 and 5 represent the percentages of the regions which the correctly recognized faces have won. But note that the percentages of the maximum vote obtained differ considerably among the candidates depending on how the votes distribute among the candidates; a high percentage of correctly recognized regions does not imply directly that the corresponding face is correctly recognized in the regional matching. Examining the data of figures 4 and 5, we may make a general statement that the percentages of the winning regions for a correctly recognized candidate are almost always higher as the number of partitions increase.
Exceptions are shown in figure 5 when we divide the nation into too many regions. This behavior is confirmed by figure 6 where the recognition rates start to fall off as the number of regions increase beyond 384 which correspond to the  region size of $5 \times 5$ pixels for the entire image size of ${80 \times 120}$ pixels. Obviously when the region size is less than $5 \times 5$ pixels, not only Assumption \ref{asp:averagedistribution} is invalidated but the eigenvector method may not give a meaningful result.
Figure 6 shows clearly that the regional matching always gives a better recognition rate than the national matching, and that the smaller the regions we choose, the higher recognition rate we will have for the regional matching.  When the regions divided are too small, say,  each region being of size less than $5 \times 5$ pixels for 384 regions and $3 \times 2$ pixels for 1600 regions in this example, the recognition rate will decrease and deteriorate. This very convincingly support the soundness of our theorems, implying also that the soundness of the Average Distribution Assumption in Section \ref{sec:theorems} and the validity of the relaxing conditions on the assumption of Section \ref{subsubsec:relaxing}. 

\begin{figure*}[hp]
\centerline{\epsfbox{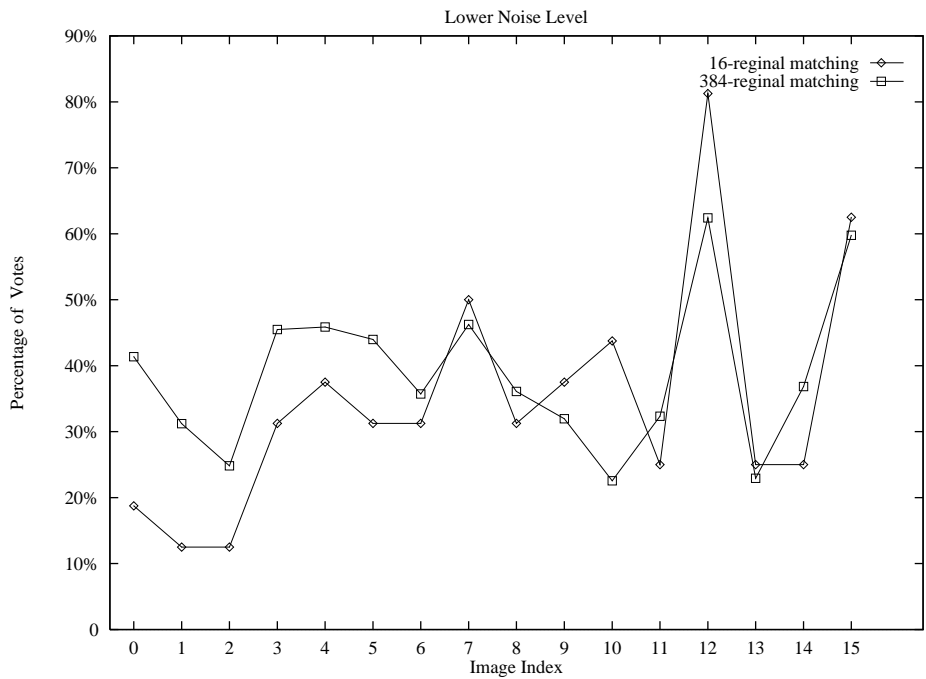}}
\caption{The Rates of Correctly Recognized Regions in Each Regional Matching Scheme for Lower Noise Level Images}
\end{figure*}

\begin{figure*}[hp]
\centerline{\epsfbox{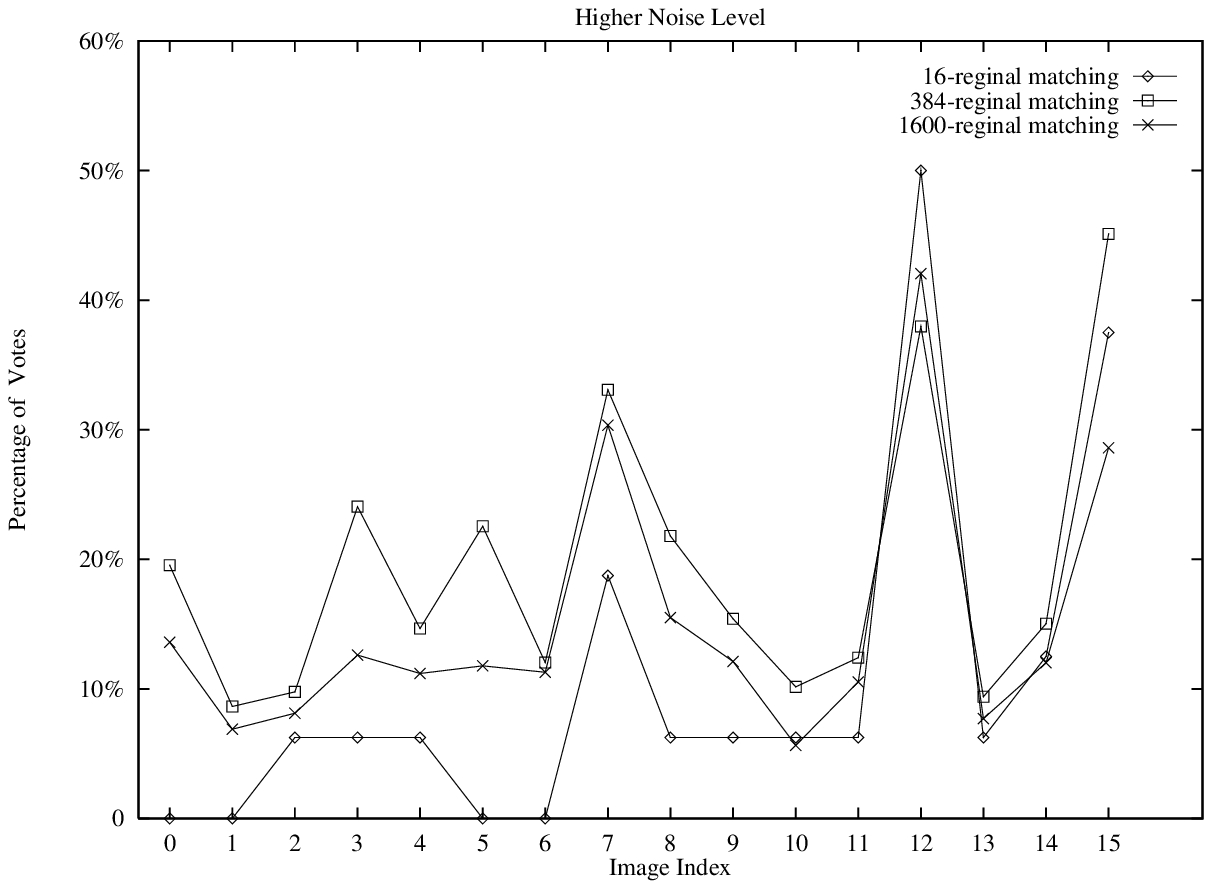}}
\caption{The Rates of Correctly Recognized Regions in Each Regional Matching Scheme for Higher Noise Level Images}
\end{figure*}

\begin{figure*}[hpt]
\centerline{\epsfbox{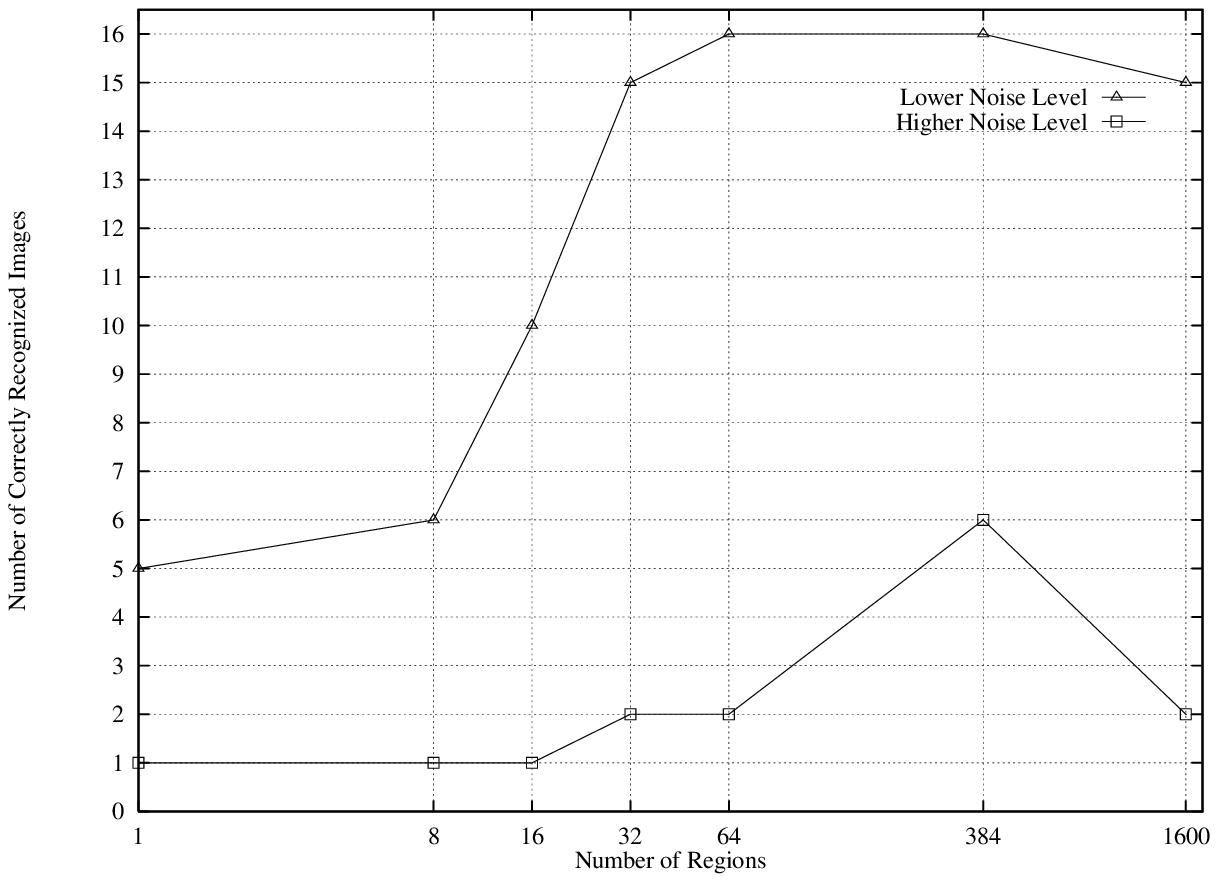}}
\caption{Recognition Rates of Each Regional Matching Scheme and National Matching}
\end{figure*}

The superiority of the regional matching is evident for images at lower noise level.  For 16 regional matching, 10 images out of 16 candidates are recognized correctly while for 384 regional partitioning, all 16 images are recognized correctly, while the Turk and Pendland's \cite{turkpentland} eigentemplate matching method on the whole image can only recognize 5 images out of 16 (figure 6)\footnote{Detailed tables are available but in the interest of space they are not given in the paper but may be downloaded from authors' homepage {\em http://alfin.mine.utsunomiya-u.ac.jp/\~\,lchen/papers/voting-tables.ps} or obtained on request to the authors.}. 
At a higher noise level, 16 regional matching recognizes only 1 images while  384 regional matching recognizes  6 (figure 6). This should be compared with the global matching of one correct recognition. Increasing the number of regions does not necessarily improve the results further. The numerical results for  1600 regions, which can recognize two, confirm this fact. This is a tradeoff problem on the size of the partitioned regions as discussed in section \ref{subsubsec:tradeoff}.

The first motivation for this work arose from the entirely same situation in facial recognition problems by Gabor wavelet analysis \cite{huangtokuda} where the matching is carried out in $8 \times 8$ Gabor regions of window (equivalent to $8 \times 8$ partitioned regions in the present paper). We are able to identify the faces 100\% using the images of 16 people under three different lighting conditions including  head-on lighting, 45 degree lighting and 90 degree lighting conditions. On the contrary, by the global vote counting method where Turk and Pentland's original eigenface algorithm \cite{turkpentland} is used, we were able to correctly identify the faces with 87.5\% accuracy. However, the comparison given there is not decisive in determining the superiority of the regional matching over the global matching.  This is so because the regional matching involved the Gabor transform as well as the eigenvector decomposition on one hand while the global matching did involve only the eigenvector decomposition. Furthermore, noise added due to different lighting conditions does not strictly satisfy the condition of localized concentration. This paper is prepared to give a solid support to the validity and stability of the wavelet-type regional matching.


\begin{thebibliography}{}

\bibitem[\protect\citeauthoryear{Aho and Ullman}{Aho and
  Ullman}{1992}]{ahoullman}
\bibsc{Aho, A.~V. and Ullman, J.~D.} \bibyear{1992}.
\newblock \bibemph{Foundations of Computer Science}.
\newblock Computer Science Press, New York.

\bibitem[\protect\citeauthoryear{Black and Anandan}{Black and
  Anandan}{1996}]{blackanandan}
\bibsc{Black, M.~J. and Anandan, P.} \bibyear{1996}.
\newblock The robust estimation of multiple motions: Parametric and
  piecewise-smooth flow fields.
\newblock \bibemphic{Computer Vision and Image Understanding}~\bibemph{63},~1,
  75--104.

\bibitem[\protect\citeauthoryear{Hochbaum and Maass}{Hochbaum and
  Maass}{1985}]{hochbaummaass}
\bibsc{Hochbaum, D.~S. and Maass, W.} \bibyear{1985}.
\newblock Approximation schemes for covering and packing problems in image
  processing and vlsi.
\newblock \bibemphic{Journal of Assocication of Computing
  Machinery}~\bibemph{32},~1, 130--136.

\bibitem[\protect\citeauthoryear{Huang, Tokuda, Miyamichi, and Chen}{Huang
  et~al.}{1999}]{huangtokuda}
\bibsc{Huang, Z., Tokuda, N., Miyamichi, J., and Chen, L.} \bibyear{1999}.
\newblock An eigenface approach to facial recognition by 2-d finite discrete
  gabor transform.
\newblock Technical report, Computer Science Department, Utsunomiya University.
\newblock Submitted to Transactions of IEICE Systems and Information.

\bibitem[\protect\citeauthoryear{Jolliffe}{Jolliffe}{1986}]{jolliffe}
\bibsc{Jolliffe, I.} \bibyear{1986}.
\newblock \bibemph{Principal Component Analysis}.
\newblock Springer Verlag, New York.

\bibitem[\protect\citeauthoryear{Lee}{Lee}{1996}]{TSLee}
\bibsc{Lee, T.~S.} \bibyear{1996}.
\newblock Image representation using 2d gabor wavelets.
\newblock \bibemphic{IEEE Trans. Pattern Analysis and Machine
  Intelligence}~\bibemph{18},~10, 959--971.

\bibitem[\protect\citeauthoryear{Pratt}{Pratt}{1991}]{pratt}
\bibsc{Pratt, W.~K.} \bibyear{1991}.
\newblock \bibemphic{Digital Image Processing} (2nd ed.).
\newblock John Wiley and Sons, Inc, New York.

\bibitem[\protect\citeauthoryear{Turk and Pentland}{Turk and
  Pentland}{1991}]{turkpentland}
\bibsc{Turk, M. and Pentland, A.} \bibyear{1991}.
\newblock Eigenfaces for recognition.
\newblock \bibemphic{Journal of Cognitive Neuroscience}~\bibemph{3},~1, 71--86.

\end{thebibliography}
\end{document}